\documentclass{article}

\usepackage{amsmath}
\usepackage{amssymb}
\usepackage{bm}
\usepackage{natbib}

\title{On Quadratic Penalties in Elastic Weight Consolidation}

\author{Ferenc Husz\'{a}r}

\usepackage[left=3cm,right=3cm]{geometry}

\begin{document}
\newcommand{\diag}{\operatorname{diag}}

  \maketitle

\begin{abstract}
Elastic weight consolidation~\citep[EWC, ][]{Kirkpatrick2017catastrophic} is a novel algorithm designed to safeguard against catastrophic forgetting in neural networks. EWC can be seen as an approximation to Laplace propagation~\citep{eskin2004laplace}, and this view is consistent with the motivation given by \citet{Kirkpatrick2017catastrophic}. In this note, I present an extended derivation that covers the case when there are more than two tasks, and show that the quadratic penalties used in EWC are not well justified and might lead to double-counting data from earlier tasks.
\end{abstract}

\section*{Introduction}
There are situations in which we would like to train a neural network to perform a range of tasks. This is usually possible if we can train the network on all tasks simultaneously. The problem is harder if we would like to train the network sequentially, one task after another. The na\:{i}ve approach of training a trained neural network on a new task via gradient descent leads to a phenomenon known as catastrophic forgetting: the network's performance in previously learned tasks rapidly deteriorates as soon as we start training on a new task.

\citet{Kirkpatrick2017catastrophic} propose a novel algorithm, elastic weight conslidation (EWC), to address this problem, while maintaining the simplicity of relying on backpropagation and stochastic gradient descent as the main algorithmic workhorses. The authors observe that catastrophic forgetting would not happen if the network's parameters were learnt in a Bayesian fashion: instead of obtaining single estimate of parameters $\theta$ via gradient descent, we calculate the Bayesian posterior distribution $p(\theta\vert \mathcal{D}_{A}, \ldots \mathcal{D}_{T})$ over possible parameter values. Here, $A\ldots T$ denote tasks and $\mathcal{D}_{A}\ldots\mathcal{D}_{T}$ the training data for each tasks, respectively. For each new task $T$ the agent learns, we simply have to update this posterior by conditioning on the new training data $\mathcal{D}_T$ and using Bayes' rule:
\begin{equation}
p(\theta\vert \mathcal{D}_{A}, \ldots \mathcal{D}_{S},  \mathcal{D}_{T}) = \frac{p(\mathcal{D}_T\vert \theta) p(\theta\vert \mathcal{D}_{A}, \ldots \mathcal{D}_{S})}{p(\mathcal{D}_T\vert \mathcal{D}_{A}, \ldots \mathcal{D}_{S})}
\end{equation}
where tasks $A\ldots S$ precede task $T$.

Due to the intractability of $p(\mathcal{D}_U\vert \mathcal{D}_{A}, \ldots \mathcal{D}_{T})$, maintaining this full posterior is not possible. However, various approximate inference techniques exist to approximate this basic procedure. EWC can be seen as an on-line approximate inference algorithm, closely related to assumed density filtering \citep[ADF][]{Opper1998ADF,minka2001expectation} and Laplace propagation \citep{eskin2004laplace}.

\section*{Laplace approximation in the two-task case}

In this section I repeat the derivation of EWC given in \citep{Kirkpatrick2017catastrophic} for the two task case, trying to keep the notation as close as possible to the original. Our goal is to approximate the Bayesian posterior over model parameters given two separate tasks $A$ and $B$. The log posterior takes the following form:
\begin{equation}
\log p(\theta\vert \mathcal{D}_A, \mathcal{D}_B) = \log p(\mathcal{D}_B\vert \theta) + \log p(\theta\vert \mathcal{D}_A) - \log p(\mathcal{D}_B\vert \mathcal{D}_A).\label{eqn:online_Bayes_twotasks}
\end{equation}

We note that Eqn.\ (2) of \citep{Kirkpatrick2017catastrophic} contains a mistake: instead of $\log p(\mathcal{D}_B\vert \mathcal{D}_A)$ the authors write $\log p(\mathcal{D}_B)$. This is irrelevant inasmuch as this term is constant with respect to $\theta$ and in EWC we never need to compute this quantity anyway.\footnote{It is tempting to believe that these two quantities are the same, i.\,e.\ $p(\mathcal{D}_B\vert \mathcal{D}_A) = p(\mathcal{D}_B)$, because data in the two tasks are drawn independently. However, dependence is induced by the Bayesian treatment of the problem: the tasks are no longer assumed independent, instead they are assumed are exchangeable (conditionally independent given $\theta$)}

The first term in the RHS of Eqn.\ \eqref{eqn:online_Bayes_twotasks}, $\log p(\mathcal{D}_B\vert \theta)$, is the log likelihood or task-specific objective function for task $B$. This is assumed to be tractable to evaluate and minimize w.r.t. $\theta$. The second term is the posterior of $\theta$ given data from the first task $\mathcal{D}_A$, which is not generally tractable.

We assume that we already trained the network to perform well on task $A$, i.\,e.\ we found a local minumum of parameters
\begin{equation}
\theta^\ast_A = \operatorname{argmin}_\theta \left\{ -\log p(\theta \vert \mathcal{D}_A) \right \}.\
\end{equation}
The gradient of $-\log p(\theta \vert \mathcal{D}_A)$ with respect to $\theta$ is $0$ at $\theta^\ast_A$, therefore $-\log p(\theta \vert \mathcal{D}_A)$ can be locally approximated as the following quadratic form (2nd order Taylor series around $\theta^\ast_A$):
\begin{equation}
  -\log p(\theta \vert \mathcal{D}_A) \approx \frac{1}{2}(\theta - \theta^\ast_A)^\top H(\theta^\ast_A) (\theta - \theta^\ast_A) + \text{constant},
\end{equation}
where $H(\theta^\ast_A)$ is the Hessian of $-\log p(\theta \vert \mathcal{D}_A)$ with respect to $\theta$, evaluated at $\theta^\ast_A$. As $\theta^\ast_A$ is assumed to be a local minimum, $H(\theta^\ast_A)$ is positive semi-definite.

Assuming that $\theta^\ast_A$ achieves near-perfect predictions on task $A$, we can approximate $H$ as
\begin{equation}
H(\theta^\ast_A) \approx N_A\cdot F(\theta^\ast_A) + H_\text{prior}(\theta^\ast_A),
\end{equation}
where $N_A$ is the number of i.\,i.\,d.\ observations in $\mathcal{D}_A$, $F(\theta^\ast_A)$ is the empirical Fisher information matrix on task $A$~\citep[see e.\,g.\ section 11 of][]{martens2014new} and $H_\text{prior}$ is the Hessian of the negative log prior $- \log p(\theta)$. As the parameter space is high dimensional, EWC makes a further diagonal approximation of $F$, treating its off-diagonal entries as $0$. These diagonal Fisher information values, denoted here by $F_{A,i}$, can be computed easily via back-propagation with minimal change to the stochastic gradient descent algorithm used to find the optimum $\theta^\ast_A$. Although in EWC the prior is ignored, here I assume that a zero-mean isometric Gaussian prior precision matrix $\lambda_{\text{prior}}I$ is used. EWC formul\ae can be recovered by considering $\lambda_{\text{prior}}=0$. This prior is consistent with the generally used practice of $L^2$ weight decay \citep[see e.\,g.\ Chapter 7 of ][]{Goodfellow2016deepbook}. Using these assumptions the approximate posterior becomes
\begin{equation}
  \log p(\theta\vert \mathcal{D}_A, \mathcal{D}_B) \approx \log p(\mathcal{D}_B\vert \theta) - \frac{1}{2}\sum_i \left(N_A F_{A,i} + \lambda_{\text{prior}}\right)(\theta_i - \theta^\ast_{A,i})^2 + \text{constant}.
\end{equation}

The above method of replacing a log posterior by a second order Taylor approximation around its maximum is known as Laplace's method \citep{mackay2003information, eskin2004laplace}, and amounts to a Gaussian approximation to the posterior around its mode. As we use a diagonal approximation to the Fisher information matrix, I will refer to EWC as a diagonal Laplace approximation. Under some regularity conditions, the Laplace approximation becomes exact in the limit of infinite data \citep{ghosal1995convergence}, but for finite data it has a tendency to underestimate the true entropy of the posterior. Therefore, sample size of each task has a non-negligible effect on the quality and behaviour of the approximation. In order to have better control over the approximation, \citet{Kirkpatrick2017catastrophic} introduce a task-specific hyper-parameter $\lambda_A$, which replaces the sample size $N_A$.

With this approximation, we can now train our network to seek the following minimum, as in EWC:
\begin{equation}
\theta^\ast_{B} = \operatorname{argmin}_\theta \left\{ - \log p(\mathcal{D}_B\vert \theta) + \frac{1}{2}\sum_i \left(\lambda_A F_{A,i} + \lambda_{\text{prior}}\right)(\theta_i - \theta^\ast_{A,i})^2 \right\}.
\end{equation}
Notice how this formula only depends on $\mathcal{D}_B$ while all data and information from task $A$ is encapsulated in the second term, a quadratic penalty that encourages $\theta$ to stay close to the previously learned parameter value $\theta_A$. We can think of this penalty as a quadratic proxy to the loss function from task $A$ which is locally correct, thereby minimizing the sum simulates the effects of training on tasks $A$ and $B$ simultaneously.

\section*{Recursive Laplace approximation for more than two tasks}

To extend EWC to the multi-task situation one can apply the diagonal Laplace approximation recursively. When learning the third task $C$, after $A$ and $B$, the Bayesian posterior decomposes as follows:
\begin{equation}
  \log p(\theta\vert \mathcal{D}_A, \mathcal{D}_B, \mathcal{D}_C) = \log p (\mathcal{D}_C\vert \theta) + \log p(\theta \vert \mathcal{D}_A, \mathcal{D}_B) + \text{constant}.
\end{equation}

We would like to approximate the intractable $\log p(\theta \vert \mathcal{D}_A, \mathcal{D}_B)$ by its second order Taylor approximation around its minimum. However, after learning task $A$, we discarded $\mathcal{D}_A$, and we no longer have access to this posterior, only to the following approximation:
\begin{align}
\log p(\theta\vert \mathcal{D}_A, \mathcal{D}_B) \approx \log p(\mathcal{D}_B\vert \theta) - \frac{1}{2}\sum_i \left(\lambda_A F_{A,i} + \lambda_{\text{prior}}\right)(\theta_i - \theta^\ast_{A,i})^2 + \text{constant}.
\end{align}

We can apply Taylor series approximation to the RHS of this formula, around its local optimum $\theta^\ast_{B}$, which we already computed when learning task $B$. Again, as $\theta^\ast_{B}$ is a local optimum, the Taylor expansion has no first order term. Notice that the quadratic penalty around $\theta^\ast_A$ has constant curvature, and its second derivative is $\lambda_A\diag(F_{A,i}) + \lambda_{\text{prior}}$. The Hessian of the negative log likelihood $- \log p(\mathcal{D}_B\vert \theta)$ can be approximated as sample size times the diagonal Fisher information $N_B \diag(F_{B,i})$ as before. Replacing $N_B$ by a hyperparameter $\lambda_B$, this gives rise to the following approximation:
\begin{equation}
  \log p(\theta\vert \mathcal{D}_A, \mathcal{D}_B, \mathcal{D}_C) \approx \log p (\mathcal{D}_C\vert \theta) - \frac{1}{2} \sum_i \left(\lambda_A F_{A,i} + \lambda_B F_{B,i} + \lambda_{\text{prior}}\right) (\theta_i - \theta^\ast_{B,i})^2 + \text{constant}.
\end{equation}

Generalizing this recursively to an arbitrarily long sequence of tasks $A,B,\ldots,S,T$ we obtain
\begin{equation}
  \theta^\ast_T = \operatorname{argmin}_\theta \left\{ - \log p (\mathcal{D}_T \vert \theta) + \frac{1}{2} \sum_i \left(\sum_{t<T} \lambda_t F_{t,i} + \lambda_{\text{prior}}\right) (\theta_i - \theta^\ast_{S,i})^2 \right\},\label{eqn:single_corrected}
\end{equation}
where the sum $\sum_{t<T}$ is over all tasks which precede $T$, $S$ is the task immediately preceding $T$ and $\theta^\ast_{S}$ is the parameter learnt for task $S$ while applying the above procedure recursively.

\section*{Contrasting with EWC penalties}

About extending EWC to multiple tasks, \citet{Kirkpatrick2017catastrophic} write ``When moving to a third task, task $C$, EWC will try to keep the network parameters close to the learned parameters of both tasks $A$ and $B$. This can be enforced either with two separate penalties or as one by noting that the sum of two quadratic penalties is itself a quadratic penalty''. Later they add: `` For each task, a penalty is added with the anchor point given by the current value of the parameters and with weights given by the Fisher information matrix times a scaling factor $\lambda$ that was optimized by hyperparameter search''.

This wording suggests that EWC uses multiple penalties, centred around mofrd of subsequent approximate posteriors $\theta^\ast_A$, $\theta^\ast_{B}, \ldots$. However, our derivation suggests that only a single penalty should be maintained, and it should be anchored at $\theta^\ast_{S}$, where $S$ is the latest task learned. It is intuitively clear why: $\theta^\ast_{B}$ was obtained while penalizing departure from $\theta^\ast_A$, therefore a penalty around $\theta^\ast_{B}$ already encapsulates the penalty around $\theta^\ast_A$. By placing a penalty around both $\theta^\ast_{A}$ and $\theta^\ast_{B}$ we are essentially double-counting the data from task $A$. In general, when applying multiple penalties as prescribed by \citep{Kirkpatrick2017catastrophic}, we introduce an unwanted systematic bias favouring tasks learned earlier on.

To a degree this systematic bias can be counterbalanced by setting $\lambda_T$ such that EWC assigns higher importance to tasks learned later. As a result, if $\lambda$ are set automatically we may never observe the order-bias in experiments.

\section*{Revisiting data from previous tasks}

The above derivation assumes a single-sweep on-line learning setting: once the optimal parameter $\theta^\ast_{A \ldots T}$ and Fisher information $F_T$ have been calculated, data from task $T$ can be thrown away. Similarly, previous parameter values $\theta^\ast_t, t<T$ can be discarded, and the Hessian $\sum_{t<T} \lambda_t F_{t,i} + \lambda_{\text{prior}}$ can be updated as a moving sum, without saving the value of $F_{t,i}$ for each task $t$. This means that the storage cost of EWC with the single penalty in Eqn.\ \eqref{eqn:single_corrected} is constant with respect to the number of tasks.

However, it may be desirable to modify the algorithm to maintain separate penalties for each task, or more precisely, each dataset $\mathcal{D}_T$. This allows us to revisit datasets to refine our Taylor approximation in light of tasks learned later\footnote{Note that there is an important difference between revisiting old data and revisiting the same task but drawing new training data}.

To obtain task-specific penalties we simply have to decompose the single penalty as a sum of penalties over tasks:
\begin{align}
\frac{1}{2} \sum_i \left(\sum_{t\leq T} \lambda_{t} F_{t,i} + \lambda_{\text{prior}}\right) (\theta_i - \theta^\ast_{T,i})^2 = \frac{1}{2} \sum_{t\leq T} \sum_i \lambda_{t} F_{t,i} (\theta_i - \tilde{\theta}_{t, i})^2 + \lambda_{\text{prior}} \sum_i \theta_i^2 + \text{constant},
\end{align} 
where $t\leq T$ denotes all tasks up to and including $T$, and the penalty centres $\tilde{\theta}_{t, i}$ can be recursively computed each step using the following formula
\begin{equation}
\tilde{\theta}_{T, i} = \frac{ \left(\sum_{t\leq T} \lambda_{t} F_{t,i} + \lambda_{\text{prior}}\right) \theta^\ast_{T,i} - \sum_{t < T} \lambda_{t} F_{t,i} \tilde{\theta}_{t,i}}{F_{T,i}}.
\end{equation}
Like $\theta^\ast_T$, this can be recursively computed for a sequence of tasks. But now, we can revisit an existing dataset by simply dropping the corresponding penalty, and treating the dataset as if it was seen for the first time. This procedure is akin to expectation-propagation \citep{minka2001expectation} or, more specifically, Laplace propagation \citep{eskin2004laplace}, while the single-sweep recursive computation is more akin to assumed density filtering \citep{Opper1998ADF}.

The individual penalties $\sum_i \lambda_{T} F_{T,i} (\theta_i - \tilde{\theta}_{T, i})^2$ can also be thought of as an approximation to the loss surface for task $T$ around $\theta^\ast_T$ - up to a constant - and can thus be used to approximately track how the agent's performance on each task evolves as new tasks are added.

\section*{Conclusions}

In the two-task case, EWC is equivalent to a diagonalized Laplace approximation, with an additional learnable task importance hyperparameter $\lambda_A$. In this note I showed that this equivalence no longer holds from the third task onwards. Instead of EWC's multiple task-specific penalties anchored at $\theta^\ast_A,\ldots \theta^\ast_T$, consistent recursive application of Laplace approximation results only a single penalty, anchored at the most recently learned parameters $\theta^\ast_T$. This makes intuitive sense, considering that $\theta^\ast_T$ was obtained while regularizing distance from previous parameters, thereby incorporating information from previous tasks. I derived a single-penalty version of EWC which should be sufficient for most cases, and whose storage costs are constant with respect to the number of tasks. I also presented a multi-penalty version of EWC with debiased task-specific penalty centres, which can be used in cases when revisiting previous datasets is a requirement.

\bibliographystyle{plainnat}
\bibliography{ewc}
 \end{document}